# An Optimized Fuzzy Logic Model for Proactive Maintenance

Abdelouadoud Kerarmi[1], Assia Kamal-idrissi[1] and Amal El Fallah Seghrouchni[1,2]

[1]Ai movement, Center of Artificial Intelligence, Mohammed VI Polytechnic University, Rabat, Morocco
[2]Lip6, Sorbonne University, Paris, France

## ABSTRACT

*Fuzzy logic has been proposed in previous studies for machine diagnosis, to overcome different drawbacks of the traditional diagnostic approaches used. Among these approaches Failure Mode and Effect Critical Analysis method(FMECA) attempts to identify potential modes and treat failures before they occur based on subjective expert judgments. Although several versions of fuzzy logic are used to improve FMECA or to replace it, since it is an extremely cost-intensive approach in terms of failure modes because it evaluates each one of them separately, these propositions have not explicitly focused on the combinatorial complexity nor justified the choice of membership functions in Fuzzy logic modeling. Within this context, we develop an optimization-based approach referred to Integrated Truth Table and Fuzzy Logic Model (ITTFLM) thats martly generates fuzzy logic rules using Truth Tables. The ITTFLM was tested on fan data collected in real-time from a plant machine. In the experiment, three types of membership functions (Triangular, Trapezoidal, and Gaussian) were used. The ITTFLM can generate outputs in 5ms, the results demonstrate that this model based on the Trapezoidal membership functions identifies the failure states with high accuracy, and its capability of dealing with large numbers of rules and thus meets the real-time constraints that usually impact user experience.*

## 1. INTRODUCTION

Fuzzy Logic (FL) is an Artificial Intelligence (AI) technique that was developed based on Fuzzy theory since it can work in the absence of data, it attempts to model and manipulate imprecise and subjective knowledge imitating human reasoning[1]. FL can be used as a knowledge model or hybrid model when data are available. The main feature of a fuzzy system is the ability to reproduce human behavior. This technique has been widely applied in the fields of intelligent control, notably in maintenance sector for fault diagnosis and prognosis.

A large body of research in the literature exists for both diagnostics and prognostics. However, many diagnosis approaches are stopping at the fault isolation step, and seldom perform fault identification. Among the diagnostic approaches, we locate Failure Mode and Effect Critical Analysis (FMECA). It is a knowledge-based approach that attempts to identify potential modes and analyze failures separately before they occur, based on experts' evaluation which is time-consuming [2]. In addition to this, this method is characterized by requiring data, and also the inability to deal with uncertain failure data including subjective expert judgments.

The use of FL-based diagnostic in literature can be classified into two groups. The first group focuses on FMECA combined with FL based on the assumptions of data certainty. The second group addresses using FL to replace FMECA but there is no analytical approach to select input members or generate rules as it is defined manually and subjectively by listing all fuzzy rules. However, none of the two groups address the question of combinatorial complexity. Therefore, in the previously published research, there are investigations about taking into consideration the combinatorial complexity while generating fuzzy rules. In the worst case, the number of generated rules corresponds to all combinations of fuzzy sets, assuming 2 input variables and 1 output variable with respectively (n, m, k) fuzzy sets, then there is $n \times m \times k$, it is the Cartesian product of fuzzy sets of all variables. To the best of author's knowledge, this is the first work focusing on studying generating, and evaluating the truth of rules swiftly.

Furthermore, most prognostic approaches assume some diagnosis has been performed and focus on the prognosis of a single failure mode. Moreover, none of these studies provides a complete framework (from data-driven diagnostic to maintenance decision passing by prognostic).

As a scope of this paper, we focus on optimizing the diagnostic step. Indeed, we developed an optimized framework to automatically generate fuzzy rules. The proposed modeling framework-referred to as Integrated Truth Table and Fuzzy Logic Model (ITTFLM) intelligently generates fuzzy logic rules using Truth Tables. This approach allows diagnosing the machine state by combining the two distinct practices in information engineering: data-driven modeling and knowledge representation. As described in Figure 1, the methodology is based, first, on data extracted from sensors, second, on the FMECA of a machine state that is selected as the knowledge source, to validate our model on real data.

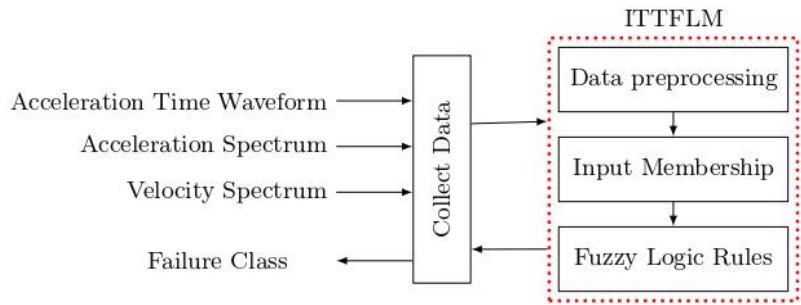

Figure 1: Three main contributions with a list of inputs and output.

In the light of the above-mentioned short comments on the aims of the study, the contributions may be shortened as follows:

1. A new reduction method to specify fuzzy sets of input memberships.
2. ITTFLM is designed to automatize FMECA processes based on Vibration data.
3. Generating rules in ITTFLM are based on Truth Table.

The remainder of this paper is organized as follows. In Section 2, we review the relevant literature. Section 3 describes the model formulation. Section 4 is dedicated to discussing the results of the experiment conducted. Finally, in Section 5 we conclude the paper and outline directions for future research.

## 2. RELATED WORK

This section first reviews the literature related to the FMECA model. Second, it reviews relevant contributions in the field of the fuzzy logic model as a background for the development of the proposed optimization model.

  a) *FMECA:* One of the main reliability analysis methods used to determine maintenance action priority. It was first developed and applied by NASA in the 1960s to improve and verify the reliability of space program hardware in the Apollo program [3]. This technique is used at diverse steps of the product life cycle in several fields, such as medical, nuclear, aerospace, and other manufacturing industries [4]. The FMECA method aims to identify potential modes and treat failures before they occur, intending to eliminate them or minimize the associated risks. It consists of systematically considering, one after the other, each component of the system studied and analyzing the causes and effects of their potential failure. Each highlighted failure is then analyzed to determine its occurrence, severity, and detect ability. The multiplication of these three values allows for calculating the criticality index, which is called the Risk Priority Number (RPN) [5]. Many authors considered FMECA and the development of risk analysis as an essential part of maintenance management strategies [6]. [7] used the FMECA approach to determine the critical equipment for maintenance in a super thermal power plant. In the military sector, it has been applied for missile equipment maintenance decisions, where it improves the efficiency and relevance of maintenance, and avoids excessive ones [8]. [9] applied the FMECA method to analyze the reliability of a metro door system. Despite its wide use, FMECA can be an issue for three main reasons [10]: (i) the subjectivity of experts' judgments to determine the three criteria of RPN, (ii) the inability to deal with uncertain failure data, (iii) the absence of analytical basis in the RPN calculation formula, and many duplicates in RPN results. To overcome these limitations, researchers have proposed models based on Fuzzy logic.

  b) *Fuzzy Logic:* This model was applied in many areas, particularly in maintenance, and has achieved good results. In the literature, we distinguish two types of fuzzy logic in diagnostic: integrated Fuzzy logic in FMECA to calculate RPN and using Fuzzy logic on vibration data in the absence of experts. [11] proposed a model in maintenance decision-making support for textile machines using vibration monitoring and vibration spectrum [1]. It also allows the utility operators to achieve precise outage predictions and optimize real-time operation and maintenance schedules for weather risk analysis in distribution outage management[12], and for scheduling predictive maintenance on communication networks [13]. Fuzzy logic proved that it is also an appropriate tool to select the maintenance strategy for a rolling mill factory [14]. The neuro-fuzzy tool ANFIS is used to evaluate the performance loss according to the degradation of components and the deviations of system input flow integrating knowledge from two different sources: expertise and real data [15],[16]used the fuzzy theory simulation method to verify FMECA results applied to determine the priority of maintenance on the equipment of a cooling system using the risk priority level method.[17] also used fuzzy logic to optimize the maintenance of power transformers, which provides a more reliable and accurate health index of transformers. Concerning FL optimization, [18] used genetic algorithms to achieve FL optimal parameters for electrical signals parameters driving based on the gaussian membership function. However, the gaussian membership function isn't adequate with the nature of vibration data intervals. Moreover, using Fuzzy logic to diagnose machines based on vibration data in maintenance applications has some limitations. On one hand, there is no method to define input members, most researchers used triangular functions without justifying this choice, on the other hand, rules are generated manually, and no algorithm guarantees the consistency and non-redundancy of rules which greatly impacts time complexity.

## 3. FUZZY MODELING

In this section, we describe the data used and then we present the framework of the proposed model (ITTFLM).

### 3.1. Data Preprocessing

The process of collecting and storing data from a physical system may result in some inconsistent, missing, or noisy values. Given that the quality of the data has a significant impact on the results achieved, it needs to be processed. First by data cleaning (filtering, transforming, removing noise). Secondly, data transformation provides a more appropriate form of data for the next step in the modeling phase. And finally, by data reduction. A large volume of data might be an issue for machine decision-making due to the high computational cost: as the volume of data rises, so will the time spent by the hardware.

The data used is uploaded from four sensors installed in an industrial facility operating in the mining industry. It contains different observations collected every 4 hours from four positions *P* for two variables fftv and fftg calculated from the acceleration time waveform *g* respectively corresponding to vibration and velocity, for approximately 251 hours. FMECA results are saved at the same time, as shown in Table 1. This data was already pre-processed and ready to be analyzed. Given that the Root Mean Square (RMS) value of velocity is one of the important factors for machinery status diagnosis, we calculated using equation 1 [19], the RMS of fftv and fftg in each sensor position for each machine state class identified by the FMECA method using the following formula:

$$x_{rms} = \sqrt{\frac{1}{n}(x_1^2 + x_2^2 + \cdots + x_n^2)} \qquad (1)$$

Table 1: Failure Class Generated by FMECA

| P[1] | g[2] | fftv[3] | fftg[4] | MSC[5] | FCC[6] |
|---|---|---|---|---|---|
| P1 – P2 – P3 – P4 | Data | Data | Data | Normal | Normal |
| | | | | Imbalance | Rotor |
| | | | | Structural fault | Frame |
| | | | | Misalignment | Link |
| | | | | Mechanical looseness | Looseness |
| | | | | Bearing lubrication | Lubrication fault |
| | | | | Gear fault | Gear |

[1] *Sensor position;* [2] *Acceleration time waveform;* [3] *Velocity spectrum.* [4] *Acceleration spectrum;* [5] *Machine State Class;* [6] *Failure Cause Class.*

### 3.2. Fuzzy Modeling

To overcome the limitations previously explained, a new methodology is developed based on Fuzzy Logic Models and the Truth Table method to simplify rules generation. Fuzzy Logic consists of four steps: initialization, fuzzification, Fuzzy rules base, defuzzification:

### 3.2.1. Initialization

**a) Intervals defining**: We started by grouping the data by machine state class. Next, we only selected the minimum and the maximum values of $[v_1,...v_n]$ and of $[g_1,...g_k]$ of each variable fftv and fftg, which allows us to create intervals $Iv = [min, max]$, and $Ig = [min, max]$ for each machine state $MSC_i$ where $i \in \{1,...,7\}$, Figure 2 represents the steps of the process.

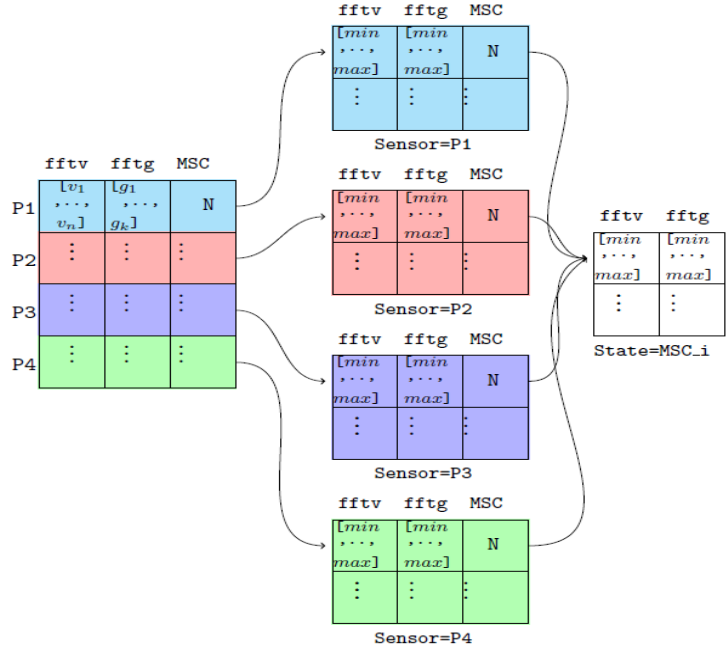

Figure 2: Data processing and intervals definition process.

As each sensor P1, P2, P3, and P4 have 1004 rows, each row contains up to 160000 observations for g, 3200 observations for fftv and fftg each. The figure above represents the process of only 4 hours of data observation (one row). First, we isolated each sensor data, one by one, then we grouped the data by state, we stated that 63.74% of the data is in a normal state, while the rest represent the 6 types of failures. This data was stored as a list in each row, which caused us a challenge during the preprocessing. We took the minimum and the maximum value of each list in a row, then, we did the same thing again for each column, this allowed us to fully cover the table and identify the minimum and the maximum value in each machine state. The final obtained table is represented in Table 2:

Table 2: Failure class of each interval

| fftv[1] | fftg[2] | MSC[3] | FCC[4] |
|---|---|---|---|
| $Iv_1$ | $Ig_1$ | Normal | Normal |
| $Iv_2$ | $Ig_2$ | Imbalance | Rotor |
| $Iv_3$ | $Ig_3$ | Structural fault | Frame |
| $Iv_4$ | $Ig_4$ | Misalignment | Link |
| $Iv_5$ | $Ig_5$ | Mechanical looseness | Looseness |
| $Iv_6$ | $Ig_6$ | Bearing lubrication | Lubrication fault |
| $Iv_7$ | $Ig_7$ | Gear | Gear fault |

[1] Velocity spectrum; [2] Acceleration spectrum.; [4] Machine State Class; [5] Failure Cause Class.

To better observe the results of the data reduction method.We plot it as a histogram. Figure 3 analyzes the relationbetween the intervals of *Iv* and *MSC* (Machine State Class).The green color corresponds to the maximum RMS of *fftv*while the purple color is the minimum.

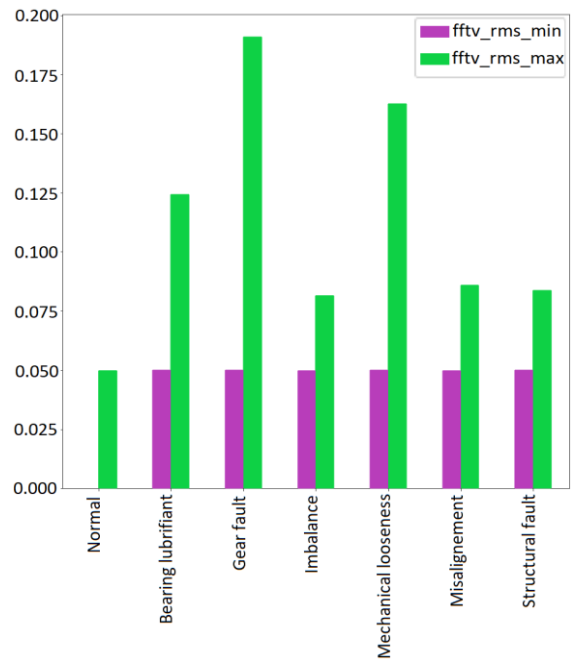

Figure 3: MSC representation according to RMS measure of fftv when generating Iv intervals.

Respectively, Figure 4 represents Ig for the metric fftg, where the orange color corresponds to the maximum RMS of fftg while the blue color is the minimum.

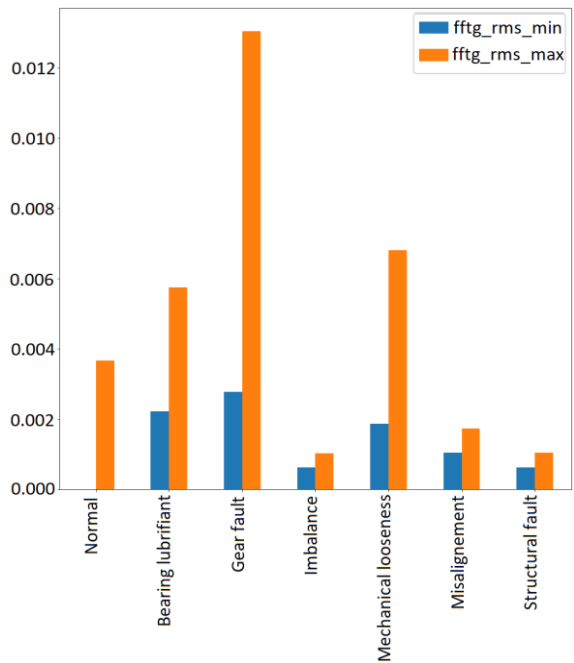

Figure 4: MSC representation according to RMS measure of fftg when generating Ig intervals.

***b) Defining machine states by intervals:*** The machinestate depends on the evolution of *Xv* and *Xg* in the intervals *Iv* and *Ig*. Applying the truth table for the two inputs variablesusing **Logical conjunction** gave us n = $2^7$ results, only 7 are possible, the table follows shows the possible results:

Table 3: Truth table for possible machine state using logical conjunction

| fftv[1] | fftg[2] | Nr[3] | Im[4] | St[5] | Mi[6] | Ml[7] | BI[8] | Gf[9] |
|---|---|---|---|---|---|---|---|---|
| $Iv_1$ | $Ig_1$ | 1 | 0 | 0 | 0 | 0 | 0 | 0 |
| $Iv_2$ | $Ig_2$ | 0 | 1 | 0 | 0 | 0 | 0 | 0 |
| $Iv_3$ | $Ig_3$ | 0 | 0 | 1 | 0 | 0 | 0 | 0 |
| $Iv_4$ | $Ig_4$ | 0 | 0 | 0 | 1 | 0 | 0 | 0 |
| $Iv_5$ | $Ig_5$ | 0 | 0 | 0 | 0 | 1 | 0 | 0 |
| $Iv_6$ | $Ig_6$ | 0 | 0 | 0 | 0 | 0 | 1 | 0 |
| $Iv_7$ | $Ig_7$ | 0 | 0 | 0 | 0 | 0 | 0 | 1 |

[1] *Velocity spectrum;* [2] *Acceleration spectrum;* [3] *Normal state.* [4] *Imbalance;* [5] *Structural fault;* [6] *Misalignment;* [7] *Mechanical looseness;* [8] *Imbalance;* [9] *Gear fault.*

A machine state is characterized depending on *fftv* and fftg values, for example, when the value Xv belongs to interval *Iv₁*and value Xg belongs to the interval *Ig₁*, the machine is in its normal state, while if *Xv* and *Xg* respectively belong to*Iv₃*and*Ig₃*, the machine will be suffering from a structural fault.

After analyzing figures 2 and 3, as well as the data in Table 3, we noticed some inclusions and intersections between intervals, considering only the inclusions for the moment, it can help with fuzzy logic rules optimization, which is a more logical and practical solution to adopt. The obtained results are represented in Table 4.

The linguistic variables correspond to the state of machines. After defining of intervals (membership functions), we checked each interval to define inclusions with other intervals, the following algorithm gives a simple way how to check the inclusion between intervals:

---

**Algorithm 1: Intervals Inclusion Detection (IIC)**
**Input: S**
**Output: S**
**1 For each two intervals A1 and A2 in S:**
$$l_1 \leftarrow length(A_1)$$
$$l_2 \leftarrow length(A_2)$$

**2 If $A_1[0] \geq A_2[0]$ and $A_1[l_1] \leq A_2[l_2]$ then:**
$$S \leftarrow S \setminus A_1$$
   End If
End For

---

The results are represented in the following table:

Table 4: Optimized truth table for each machine state

| fftv[1] | fftg[2] | Nr[3] | Im[4] | St[5] | Mi[6] | Ml[7] | Bl[8] | Gf[9] |
|---|---|---|---|---|---|---|---|---|
| $Iv_1$ | $Ig_1$ | 1 | 0 | 0 | 0 | 0 | 0 | 0 |
| $Iv_2$ | $Ig_1$ | 0 | 1 | 0 | 0 | 0 | 0 | 0 |
| $Iv_4$ | $Ig_1$ | 0 | 0 | 1 | 0 | 0 | 0 | 0 |
| $Iv_5$ | $Ig_5$ | 0 | 0 | 0 | 1 | 0 | 0 | 0 |
| $Iv_7$ | $Ig_5$ | 0 | 0 | 0 | 0 | 1 | 0 | 0 |

[1] Velocity spectrum; [2] Acceleration spectrum; [3] Normal state. [4] Imbalance; [5] Structural fault; [6] Misalignment; [7] Mechanical looseness; [8] Imbalance; [9] Gear fault.

### 3.2.2. Fuzzy Rules Base:

Intervals inclusion can help with optimizing FL rules. Given that we are aiming to generate more than only one output, our model can generate possible machine states in real-time andgives the decision-making step for the agents. This will minimize time, costs, and resources. Moreover, it is an important factor to not eliminate the human factor, there will be a collaboration between Human and machine capabilities. In our algorithm, each rule is generated as a combination of the degree of each input and output variable a teach step. Each row of the truth table represents a rule of FuzzyInference, it contains one possible configuration of the input and output variables in the table according to linguistic terms defined for each variable, which are machine state. The idea is to optimize the generation process by ensuring complete and fast fuzzy rules based on logical evaluation rather than the linguistic rule. To the best of our authors knowledge, this is the initial attempt that merges Truth Tables and FL.

The fuzzy rules base consisted of 7 optimized rules based on Table 4, as follows:

Table 5: Fuzzy Logic rules

| Rule | | Iv[1] | | Ig[2] | | MSC[3] |
|---|---|---|---|---|---|---|
| 1 | | $Iv_1$ | | $Ig_1$ | | Normal |
| 2 | | $Iv_2$ | | $Ig_1$ | | Imbalance |
| 3 | if | $Iv_4$ | And | $Ig_1$ | then | Structural fault |
| 4 | | $Iv_4$ | | $Ig_1$ | | Misalignment |
| 5 | | $Iv_5$ | | $Ig_5$ | | Mechanical looseness |
| 6 | | $Iv_7$ | | $Ig_5$ | | Bearing lubrication |
| 7 | | $Iv_7$ | | $Ig_5$ | | Gear fault |

[1]Intervals of fftv; [2] Intervals of fftg; [3] Machine State Class.

### 3.2.3. Defuzzification

In this phase, the system examines all of the rule outcomes after they have been logically added and then computes the final output value of the fuzzy controller. The results are shown in Section4.

## 4. CASE STUDY

Our experiment protocol aims to answer the following questions:
   a) Can the inclusions and intersections impact the outputs of the fuzzy logic controller?
   b) How can we generate optimized fuzzy logic rules intelligently?
   c) Which one of the Fuzzy logic sets is the best for our case?

### 4.1. Experimentation Environment

Our model has been developed with **Python 3.9** using the **skfuzzy library**, on **Jupiter Notebook v6.4.5**, in **Anaconda Navigator v2.1.1**.

The Mamdani Fuzzy Controller is selected, and two fuzzy inputs are added to the model. The inputs are Xv and Xg.

   a) *Input membership function:* Table 5 helps in integrating linguistic terms in the identification of the inputs memberships (Fuzzification), for each system, the output is represented in Figure 5. Three fuzzy decision systems were built, using the trapezoidal, triangular, and Gaussian membership functions.

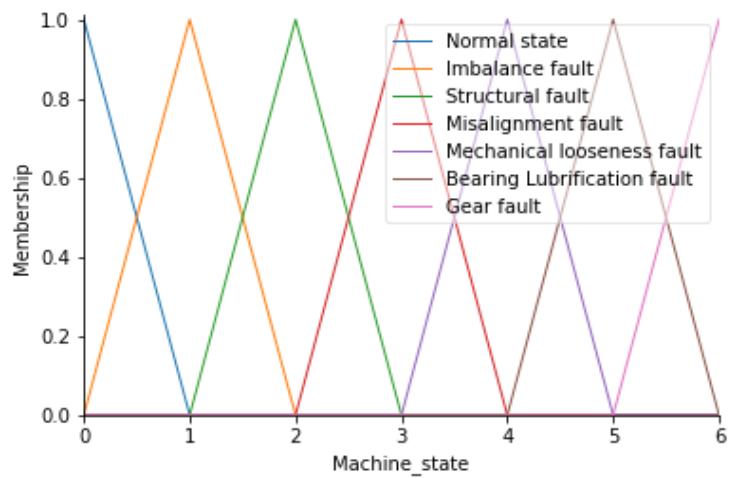

Figure 5: The output linguistic variables– Defuzzification.

   b) *Fuzzy rules set:* The fuzzy rules are presented in Table 5.

### 4.2. Experimentation Results

We ran three experiments in order to compare the three types of fuzzy logic sets, the triangular membership functions (triMF), Gaussian membership functions (gaussMF), and trapezoidal membership functions (trapMF). We set two variables from each interval, one is near the minimum value, and the other is near the maximum value of each interval. vmin for the minimum value of one of Iv's intervals, vmax for the maximum value of one of Iv's intervals, gmin for the minimum value of one of Ig's intervals, and gmax for the maximum value of one of Ig's intervals.

The obtained results of each system are presented next, and compared to the real data used data, where Exc corresponds to Excellent and Ave to Average.

**4.2.1. Gaussian Membership Functions (GaussMF):**

For the Gaussian membership functions (gaussMF), the following figure represents the input membership of Iv:

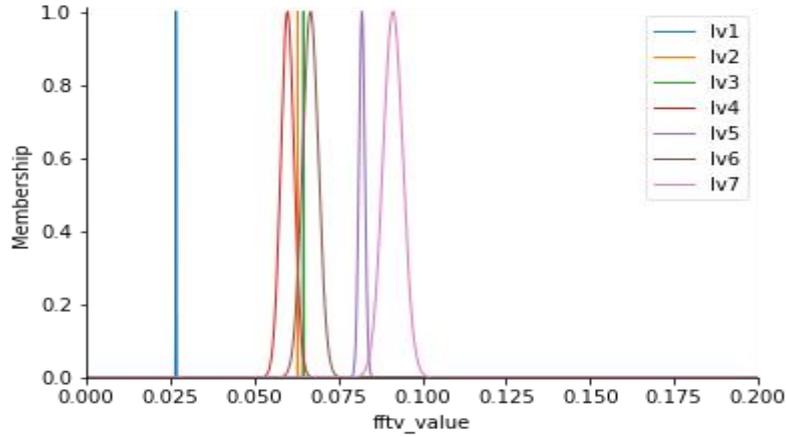

Figure 6: GaussMF - fftv input membership.

We chose twovariables Xv and Xg belonging to Iv4 and Ig4 where Machine State Class is Structural fault totest the model. The following figure shows the outputs of the system:

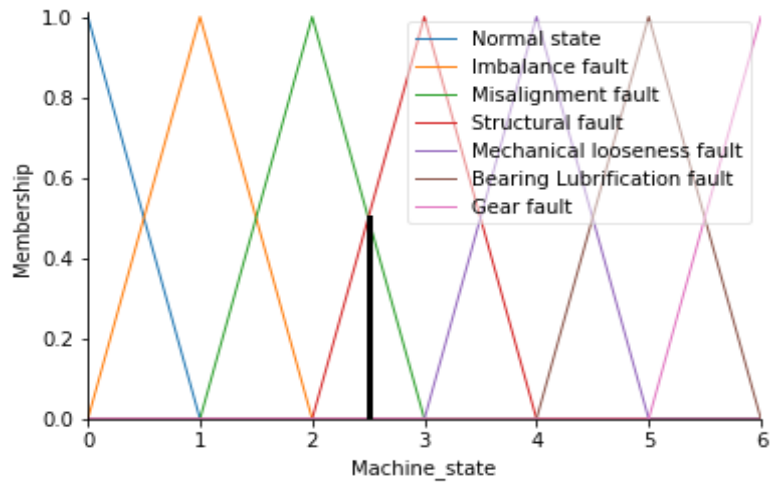

Figure 7: GaussMF - Structural fault output

The obtained results are on average and corresponded to the same given Machine State Class. The complete experiment results are presented in Table 6.

Table 6: GaussMF - Results of the experiment

| N° | fftv | fftg | ExpS[1] | GaussSc[2] | GaussS[3] | GaussA[4] |
|---|---|---|---|---|---|---|
| 1 | $v_{1-min}$ | $g_{1-min}$ | Nr | NaN | NaN | Bad |
| 2 | $v_{1-max}$ | $g_{1-max}$ | Nr | NaN | NaN | Bad |
| 3 | $v_{2-min}$ | $g_{2-min}$ | Im | 2.50 | St 50% & Mi 50% | Ave |
| 4 | $v_{2-max}$ | $g_{2-max}$ | Im | 2.50 | St 50% & Mi 50% | Ave |
| 5 | $v_{3\text{-min}}$ | $g_{3\text{-min}}$ | St | 2.50 | St 50% & Mi 50% | Ave |
| 6 | $v_{3\text{-max}}$ | $g_{3\text{-max}}$ | St | 2.50 | St 50% & Mi 50% | Ave |
| 7 | $v_{4\text{-min}}$ | $g_{4\text{-min}}$ | Mi | 0.00 | Nr 100% | Bad |
| 8 | $v_{4\text{-max}}$ | $g_{4\text{-max}}$ | Mi | NaN | NaN | Bad |
| 9 | $v_{5\text{-min}}$ | $g_{5\text{-min}}$ | Ml | NaN | NaN | Bad |
| 10 | $v_{5\text{-max}}$ | $g_{5\text{-max}}$ | Ml | 0.00 | Nr 100% | Bad |
| 11 | $v_{6\text{-min}}$ | $g_{6\text{-min}}$ | Bl | NaN | NaN | Bad |
| 12 | $v_{6\text{-max}}$ | $g_{6\text{-max}}$ | Bl | 1.76 | St 75% & Im 25% | Bad |
| 13 | $v_{7\text{-min}}$ | $g_{7\text{-min}}$ | Gf | NaN | NaN | Bad |
| 14 | $v_{7\text{-max}}$ | $g_{7\text{-max}}$ | Gf | 0.00 | Nr 100% | Bad |

[1]*State by Experts;* [2] *Score of Gaussian;* [3]*State of Gaussian.* [4]*Gaussian Accuracy;*

For the Gaussian membership functions (gaussMF), many outputs couldn't be generated, while the obtained ones show a lack of accuracy in identifying the machine state. The total of obtained results with average accuracy presents only 28% of the total tests, this might be due to the nature of the data we have and the fact that the input values using this type made the functionsnot connected which makes the system is too sparse as Figure 5 shows.

### 4.2.2. Triangular Membership Functions (triMF)

In the triangular membership functions (triMF) set, the input memberships of Iv are presented in the following figure:

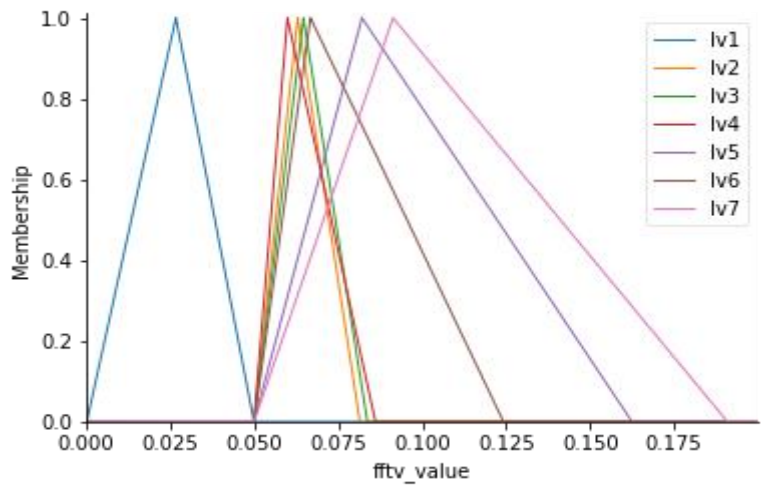

Figure 8: TriMF - fftv input membership.

In the first test for the triMF, the two variables Xv and Xg used are belonging to Iv2 and Ig2 where Machine State Class is Imbalance. The following figure shows the outputs of the system:

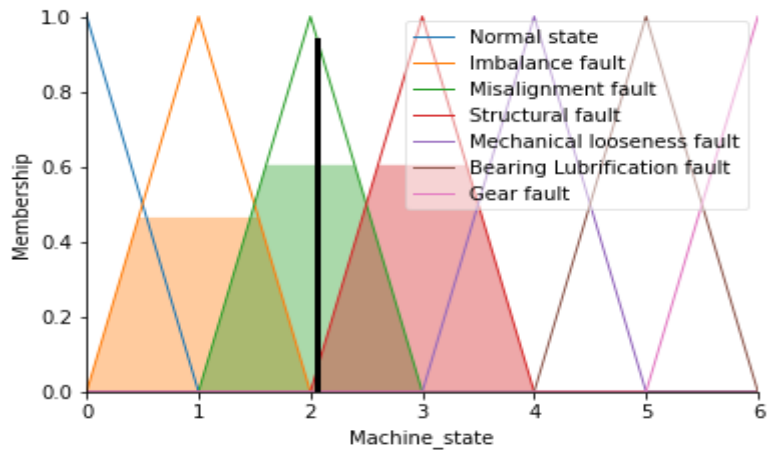

Figure 9: TriMF - Imbalance fault output.

As a result of the first test, the model gave the wrong machine state class, The completeexperiment results are presented in Table 7.

The triangular membership functions (triMF), could identify with good accuracy many machine states, but not accurate in some others, this is due to the fact that this type of fuzzy logic set is considering the median of the value as the point where a state is 100% exist as in Figure 8, while the values tested are near the borders of each interval, moreover, the tested data is a set of values that varies anywhere between the borders, not necessarily in the middle, which is more realistic. Considering the intersections between intervals, the outputs of this model gave states that correspond to other intervals, for the 4[th] test and the 7[th]among others, the model poorly defined the correct state, but not wrong. This model could detect correctly 50% of the tests. Therefore, should be checked by experts which will take more time.

Table 7: TriMF - Results of the experiment

| N° | fftv | fftg | ExpS[1] | TriSc[2] | TriS[3] | TriA[4] |
|---|---|---|---|---|---|---|
| 1 | $v_{1-min}$ | $g_{1-min}$ | Nr | 0.49 | Nr 70% & Im 30% | Good |
| 2 | $v_{1-max}$ | $g_{1-max}$ | Nr | 0.40 | Nr 60% & Im 40% | Good |
| 3 | $v_{2-min}$ | $g_{2-min}$ | Im | 2.06 | St 95% & Mi 5% | Poor |
| 4 | $v_{2-max}$ | $g_{2-max}$ | Im | 2.02 | St 100% | Poor |
| 5 | $v_{3-min}$ | $g_{3-min}$ | St | 2.06 | St 95% & Mi 5% | Exc |
| 6 | $v_{3-max}$ | $g_{3-max}$ | St | 2.04 | St 95% & Mi 5% | Exc |
| 7 | $v_{4-min}$ | $g_{4-min}$ | Mi | 2.00 | St 100% | Poor |
| 8 | $v_{4-max}$ | $g_{4-max}$ | Mi | 2.07 | St 95% & Mi 5% | Poor |
| 9 | $v_{5-min}$ | $g_{5-min}$ | Ml | 2.38 | St 65% & Mi 35% | Bad |
| 10 | $v_{5-max}$ | $g_{5-max}$ | Ml | 4.64 | Ml 65% & Bl 35% | Good |
| 11 | $v_{6-min}$ | $g_{6-min}$ | Bl | 3.11 | Mi 90% & Ml 10% | Poor |
| 12 | $v_{6-max}$ | $g_{6-max}$ | Bl | 4.59 | Ml 60% & Bl 40% | Good |
| 13 | $v_{7-min}$ | $g_{7-min}$ | Gf | 2.98 | Mi 100% | Poor |
| 14 | $v_{7-max}$ | $g_{7-max}$ | Gf | 5.51 | Gf 50% & Bl 50% | Good |

[1]State by Experts; [2] Score of Triangular; [3]State of Triangular. [4]TriangularAccuracy;

### 4.2.3. Trapezoidal Membership Functions (TrapMF)

The following figure represents the input membership of *Iv* for the trapezoidal membership functions (trapMF) set:

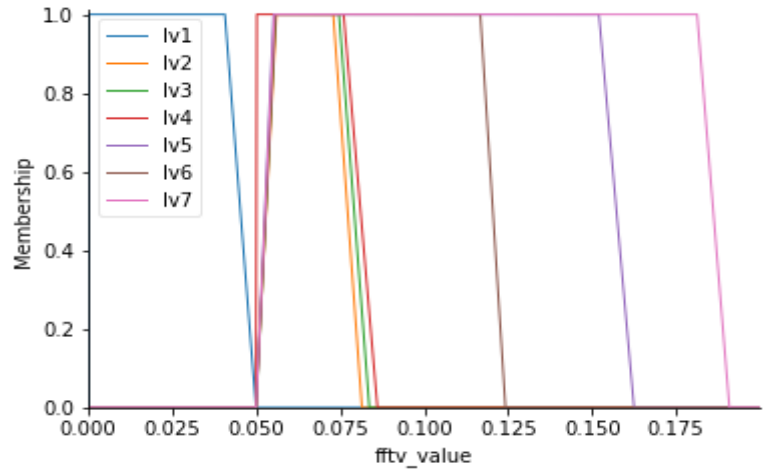

Figure 10: TrapMF - fftv input membership.

The same as previous tests we have inserted two variables *Xv* and *Xg* belonging to *Iv₁* and *Ig₁* where Machine State Class is Normal to test the model. The following figure shows the outputs of the system:

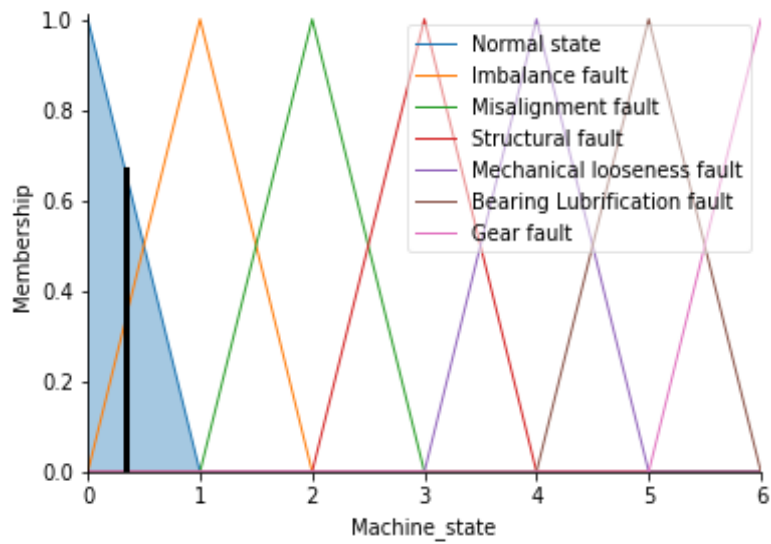

Figure 11: TrapMF - Normal state output.

The first result is obtained with good accuracy of 70%, the complete experiment results are presented in Table 8.

Table 8: TrapMF - Results of the experiment

| N° | fftv | fftg | ExpS[1] | TrapSc[2] | TrapS[3] | TrapA[4] |
|---|---|---|---|---|---|---|
| 1  | $v_{1-min}$ | $g_{1-min}$ | Nr | 0.34 | Nr 70% & Im 30% | Good |
| 2  | $v_{1-max}$ | $g_{1-max}$ | Nr | 0.33 | Nr 70% & Im 30% | Good |
| 3  | $v_{2-min}$ | $g_{2-min}$ | Im | 2.00 | St 100% | Poor |
| 4  | $v_{2-max}$ | $g_{2-max}$ | Im | 2.00 | St 100% | Poor |
| 5  | $v_{3-min}$ | $g_{3-min}$ | St | 2.00 | St 100% | Exc |
| 6  | $v_{3-max}$ | $g_{3-max}$ | St | 2.03 | St 98% & Mi 2% | Exc |
| 7  | $v_{4-min}$ | $g_{4-min}$ | Mi | 2.46 | St 55% & Mi 45% | Good |
| 8  | $v_{4-max}$ | $g_{4-max}$ | Mi | 2.08 | St 90% & Mi 10% | Ave |
| 9  | $v_{5-min}$ | $g_{5-min}$ | Ml | 3.02 | Mi 98% & Ml 2% | Poor |
| 10 | $v_{5-max}$ | $g_{5-max}$ | Ml | 4.73 | Bl 75% & Ml 25% | Ave |
| 11 | $v_{6-min}$ | $g_{6-min}$ | Bl | 3.96 | Ml 100% | Poor |
| 12 | $v_{6-max}$ | $g_{6-max}$ | Bl | 4.73 | Bl 75% & Ml 25% | Good |
| 13 | $v_{7-min}$ | $g_{7-min}$ | Gf | 3.90 | Ml 90% & Mi 10% | Poor |
| 14 | $v_{7-max}$ | $g_{7-max}$ | Gf | 5.66 | Gf 70% & Bl 30% | Good |

[1]State by Experts; [2] Score of Trapezoidal; [3]State of Trapezoidal. [4]Trapezoidal Accuracy;

The trapezoidal membership functions (trapMF) gave the best results compared to GaussMF and triMF models, this is due to the ability of the TrapMF to covera wide range of intervals, as in Figure 8, the results are also more accurate, in some cases we considered the accuracy poor due to the inclusion between intervals, in fact, the values that are near the maximum value in an interval gave excellent accuracy, while that the maximum value of each interval can be included in another one, which explains the poor accurate obtained results but not wrong. Noting that 78.57% of the results are correct and accurate, this model will provide recommendations for the maintenance team and save them maintenance intervention time.

The previous tables summarize the obtained results of the tested models, the trapMF model gave the best results compared to the other tested models, as mentioned before, our model as a first objective is to accurately define machine different health states in real-time, based on experts' knowledge and vibration data. Furthermore, this model unlike previously mentioned publications, it considered also the optimization of fuzzy rules, which help in reducing the subjectivity in putting fuzzy rules.

## 5. CONCLUSION & FUTURE WORKS

In this paper, we propose an *ITTFLM* to generate smartly FL rules based on Truth Tables. Moreover, we propose to justify the choice of membership function by simulation method. In terms of business context, this study has achieved two major goals. The first one is that it proves that it is possible to conserve old FMECA results and used them as references in real-time diagnostics. the second achievement is combining experts' knowledge with numerical data using AI, which gave more accurate and reliable results that will minimize the time of all interventions, it takes only 5ms to diagnose machines states. This model allows the agents to go straight to the source of the problem and solve it, in a short time, with fewer resources, and avoid failures that would stop the entire production process. The obtained results show that the trapezoidal membership functions (trapMF) gave the best results, and better accuracy compared to the other sets, it can be explained by the fact that it gives a wide range in which a variable can belong, also it's more realistic and practical for our case. In fact, the intersections and the inclusion between intervals can impact the model's accuracy, the results show that the Fuzzy logic technique can reduce this issue by limiting the probable cases and giving more accurate results. Later, we aim to evaluate the robustness of our model by including more data (metrics and observations) and also comparing it to other models such as FNN (Fuzzy Neural Network). The next step will be machine state prognostic based on the result of the diagnostic step. This will not only the upcoming failure but a detailed machine state.

## ACKNOWLEDGMENTS

The authors would like to express their appreciation to Dr. Btissam El Khamlichi of Mohammed VI Polytechnic University for her diligent proofreading of this manuscript. The authors are also grateful to the OCP-Maintenance Solutions (OCP-MS), the subsidiary of the Office Chérifien des Phosphates in The Morocco Kingdom (OCP group).